\documentclass[journal]{IEEEtran}
\usepackage{fixltx2e}
\usepackage[usenames,dvipsnames]{color}
\usepackage[usenames,dvipsnames]{xcolor}
\usepackage{graphicx}
\usepackage{amsmath}
\usepackage{algorithmic}
\usepackage{array}
\usepackage{enumitem}
\usepackage{caption}
\usepackage{soul}

\newcommand{\noopsort}[1]{}

\colorlet{tmcol}{Peach!20}
\colorlet{gmcol}{Emerald!20}
\colorlet{lccol}{Lavender!20}

\newlist{steps}{enumerate}{1}
\setlist[steps,1]{
  label={(\arabic*)},
  leftmargin=*,
  align=left,
  labelsep=3mm
}

\graphicspath{{./img/final_figures/vector/smaller/}}

\begin{document}
%
\title{A Gamut Mapping Framework\\for Color-Accurate Reproduction of HDR Images}
%
%

\author{\IEEEauthorblockN{Elena Sikudova\IEEEauthorrefmark{1},
Tania Pouli\IEEEauthorrefmark{2},
Alessandro Artusi\IEEEauthorrefmark{3},
Ahmet O\u{g}uz Aky\"{u}z\IEEEauthorrefmark{4},\\ 
Francesco Banterle\IEEEauthorrefmark{5}, 
Zeynep Miray Mazlumoglu\IEEEauthorrefmark{4}, and
Erik Reinhard\IEEEauthorrefmark{2}}
\thanks{\IEEEauthorblockA{Corresponding author. \IEEEauthorrefmark{1}Department of Applied Informatics, Comenius University Bratislava, Slovakia. E-mail: {sikudova@sccg.sk}.}}
\thanks{\IEEEauthorblockA{\IEEEauthorrefmark{2}Technicolor Research \& Innovation, Rennes, France}}
\thanks{\IEEEauthorblockA{\IEEEauthorrefmark{3}Department of Computer Science, Applied Mathematics and Statistics, (GiLab) University of Girona, Spain}}
\thanks{\IEEEauthorblockA{\IEEEauthorrefmark{4}Middle East Technical University, Ankara, Turkey}}
\thanks{\IEEEauthorblockA{\IEEEauthorrefmark{5}Visual Computing Laboratory, ISTI-CNR, Pisa, Italy}}}

\markboth{IEEE Computer Graphics and Applications}%
{Shell \MakeLowercase{\textit{et al.}}: Bare Demo of IEEEtran.cls for Journals}
%

\IEEEpubid{ \makebox[\columnwidth]{DOI: 10.1109/MCG.2015.116 \~\copyright 2016 IEEE. \hfill} \hspace{\columnsep}\makebox[\columnwidth]{ } }


\maketitle

\begin{abstract}
Few tone mapping operators (TMOs) take color management into consideration, limiting compression to luminance values only. This may lead to changes in image chroma and hues which are typically managed with a post-processing step. However, current post-processing techniques for tone reproduction do not explicitly consider the target display gamut. Gamut mapping on the other hand, deals with mapping images from one color gamut to another, usually smaller, gamut but has traditionally focused on smaller scale, chromatic changes. In this context, we present a novel gamut and tone management framework for color-accurate reproduction of high dynamic range (HDR) images, which is conceptually and computationally simple, parameter-free, and compatible with existing TMOs. In the CIE $LCh$ color space, we compress chroma to fit the gamut of the output color space. This prevents hue and luminance shifts while taking gamut boundaries into consideration. We also propose a compatible lightness compression scheme that minimizes the number of color space conversions. Our results show that our gamut management method effectively compresses the chroma of tone mapped images, respecting the target gamut and without reducing image quality.
\end{abstract}


%
\IEEEpeerreviewmaketitle

\section{Introduction}
\label{sec:Introduction}

High dynamic range (HDR) imaging consists of tools and 
techniques to capture, store, transmit and display images with significantly 
higher fidelity than can be achieved with conventional imaging techniques. An
important aspect of HDR imaging involves the reproduction of images on conventional displays.
As in this case the dynamic range of the image can be much higher than the
display device can accommodate, dynamic range reduction techniques need to be
employed~\cite{Rein2010,Ba+11}.

In many cases, tone reproduction techniques focus on range compression
along the luminance dimension, either leaving chromaticities
unaltered, or treating color management as a separate
problem~\cite{Sc94,Ma+09,Pouli:2013}. In the latter case, algorithms
focus on correcting or improving the appearance of the tonemapped
image. Some HDR color appearance models do integrate color and
luminance management, for the purpose of predicting the human visual
response to a stimulus \cite{Kuang2007,Reinhard:2012}. 
Such algorithms can be used successfully as display algorithms, albeit 
still without appropriate gamut management.


To our knowledge, none of the existing algorithms take the target
color gamut into consideration and as a result often produce pixel
values that cannot be correctly represented or displayed. Our method
aims to combine the color correction step often necessary after tone mapping
with gamut management into an integrated HDR gamut management
framework that handles both lightness and chroma compression, while
limiting as much as possible hue shifts and luminance distortion.
\begin{figure*}[t]
\centerline{
 \includegraphics[width=\textwidth]{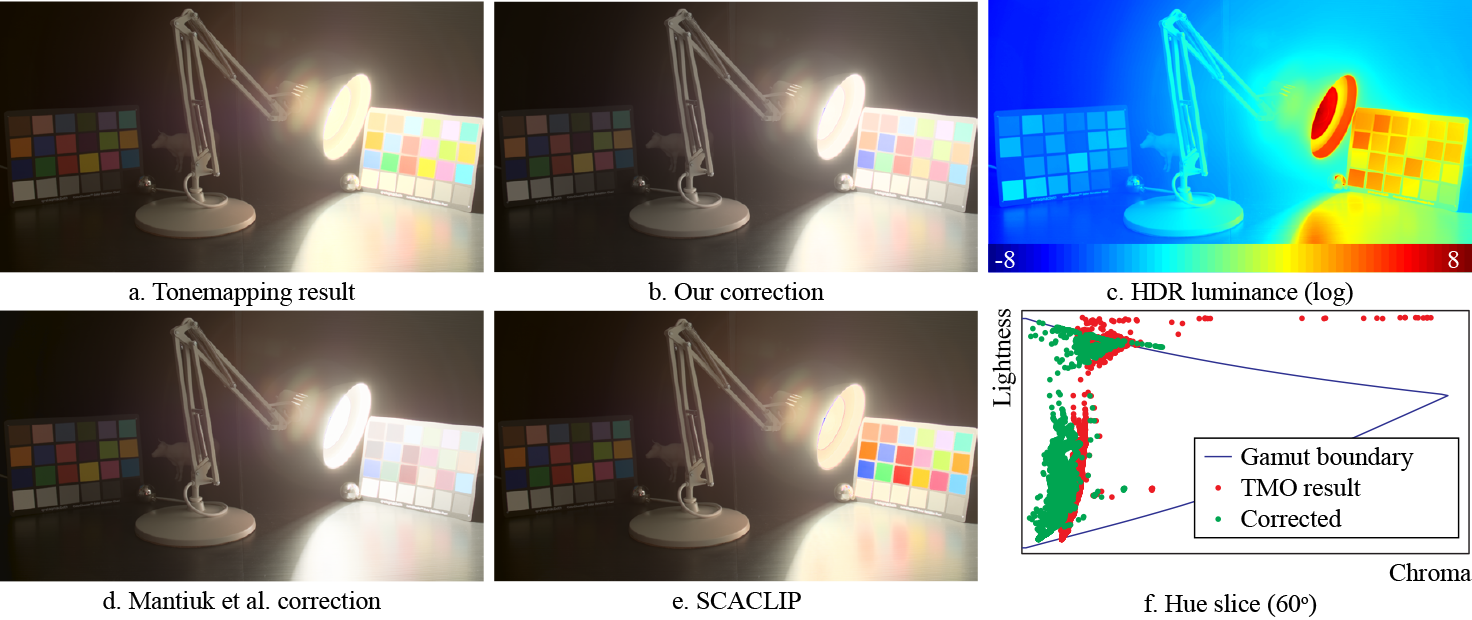}
}
   \caption{An HDR image, shown in (c) in false-color, was tone mapped using the Photographic
       operator~\cite{Rein2002a} (a) and processed with our
           framework (b), a color correction solution~\cite{Ma+09} (d) and a gamut mapping solution~\cite{Bo+07} (e). Although the tone mapping process has successfully compressed the luminance of the image, it has led to an over-saturated appearance and out-of-gamut pixels (f). Our method corrects both issues, while alternative solutions only handle one of the two.}
\label{fig:Teaser}
\end{figure*}

There are good reasons why these two dimensions should be treated in
conjunction. First, human vision does not treat colorfulness separately from
luminance values, as evidenced by the Hunt effect. In essence,
lighter objects are seen as more colorful, and vice-versa. In the context of
dynamic range management, this means that compression of luminances should be
accompanied by a corresponding adjustment of chroma. Additionally, current consumer display trends are pushing towards a combined increase in gamut and dynamic range, necessitating algorithms that can manage both of these aspects in conjunction.

Second, color spaces are three dimensional, bounded by their
gamut~\cite{Morovic2008}. The gamut spanned by the pixels in an image may not be
matched to that of the target display. In such cases, gamut mapping involves
compensating for differences in size, shape and location between the image and
display gamuts. This constitutes a mapping from one three-dimensional shape to
another. As the overlap between gamuts is typically large in conventional gamut
mapping scenarios, such algorithms aim to find a trade-off between moving
out-of-gamut pixels inside the target gamut, while pixels that are already
inside the target gamut are left alone as much as possible.

On the other hand, one-dimensional luminance adjustment, as achieved by many
tone reproduction operators, may create pixels that lie outside the target display's gamut (see Figure~\ref{fig:Teaser}f) along the chroma channel. 
These pixels are either typically not managed or treated with na\"{\i}ve approaches~\cite{Sc94} 
that tend to over-saturate as shown in Figure~\ref{fig:Teaser}a.  

Our novel gamut mapping framework integrates tone mapping with gamut
mapping, correcting the colors after dynamic range
compression, while ensuring that images fit within the target
gamut. It does not require calibrated input and is parameter-free. We
accomplish this by transforming the original HDR image into the CIE $L^*C^*h^*$
 color space, and then compressing both the lightness and chroma
channels as customary in gamut mapping algorithms. The lightness
channel can be compressed with any existing tone mapping operator, or
a scheme similar to our chroma compression can be applied to lightness
as well.

\begin{figure*}
\fcolorbox{black}{gmcol}{
\begin{minipage}[t]{\textwidth}
\section*{Side-bar: Gamut Mapping}

Given a color space, a gamut for a device or medium can be thought of as a
subspace within the color space, containing the colors that can be reproduced by
that device. A new color that is outside the gamut, cannot be accurately
reproduced by the device. To define how colors outside this gamut should be
treated or mapped to the displayable color subset, gamut mapping techniques may
be employed.

Gamut mapping techniques can be categorized as \emph{global} and \emph{spatial}.
Global techniques can be further classified as \emph{clipping} and
\emph{compression} based approaches. Clipping only changes the colors that are
located outside of the destination gamut by clamping them to the
boundaries of the destination gamut~\cite{Morovic2008}. Although it
has the advantage of preserving within-gamut colors, it is only a
viable solution if the difference between the two gamuts is
small. Compression, on the other hand, changes all the colors of the
input gamut to be adjusted into the destination
gamut~\cite{Morovic2008}. Different types of compression functions have
been proposed such as \emph{linear}, \emph{piecewise linear}, and
\emph{sigmoidal}. Compression is typically performed on both
lightness and chroma components. In contrast to the global approaches,
spatial gamut mapping attempts to preserve local information. These
methods will map similar out-of-gamut colors to the same color if they
are spatially distant in the image, but to distinct colors if they
share an edge.

Our work aims to extend existing work on gamut mapping for low dynamic
range, proposing a gamut mapping management framework to work directly
with HDR input data. This can be either integrated into existing TMOs
or be a fully stand alone solution with its own lightness compression
technique for (HDR) luminance values.

\end{minipage}
}%
\end{figure*}

\section{Related Work}
\label{ScRelated}

Reproduction of visual content on devices of different gamuts is typically
divided into two categories, namely \emph{gamut mapping} and \emph{tone
mapping}. Traditionally the first class of techniques deals with mapping the
color gamut between devices, attempting to produce the most accurate
reproduction of colors possible given the restrictions of a given device or
medium~\cite{Morovic2008}. Tone mapping, on the other hand, is primarily
concerned with compressing the luminance range of an HDR image or video such
that the media can be visualized on a low dynamic range (LDR) display
device~\cite{Rein2010,Ba+11}.

In contrast to tone or
gamut mapping, color appearance models (CAMs) predict human visual perception 
of patches of color and images. They consider parameters
relating to the scene and viewing environment and as such require
accurate measurements as input. These methods can
accurately reproduce the appearance of an image in different devices
and viewing conditions, but do not take gamut boundary issues into
consideration.

Although both tone mapping and gamut mapping research aim to reproduce
images on devices of more limited capabilities, they have remained largely disconnected areas. 
In this work, we bring together these two fields (see side-bars) in a novel gamut management
technique that can successfully compress the chromaticities of high
dynamic range (HDR) images so that they correctly match the
compression applied by any given tone mapping operator. 

\begin{figure*}
\fcolorbox{black}{tmcol}{
\begin{minipage}[t]{\textwidth}
\section*{Side-bar: Tone Mapping}

Tone mapping is typically used to prepare HDR images for display on
LDR display
devices. While tone mapping can be considered a form of gamut mapping, there are
important differences. First, tone mapping is generally employed when
the dynamic range of the input image is vastly higher than the dynamic
range of the display device. Second, tone mapping is generally
concerned with compressing luminances, while gamut mapping is
concerned with compressing perceptual attributes of lightness and
chroma. As such, it is possible that a tone mapped image will contain
out-of-gamut colors, which are clipped to gamut boundaries in an
uncontrolled manner.

An additional concern when tone mapping the luminance channel only is
that images tend to acquire an over-saturated appearance either
globally or locally, as shown in Figure~\ref{fig:Teaser}a. Appearance
aspects such as saturation and colorfulness of an image or image patch
depend both on the chromatic information and the image
luminance. Thus, to fix these color distortions, most TMOs are
augmented with a post processing step that desaturates the image by
means of a manually controlled parameter~\cite{Sc94}, while
psychophysical studies have linked this saturation parameter to the
amount of contrast correction computed from the global tone mapping
curve~\cite{Ma+09}.  Alternatively, the amount of (de-)saturation can
be computed by comparing the original HDR input to the tone mapped
result~\cite{Pouli:2013}. Although these methods can improve the
appearance of the tone mapped image, they are not able to consider the
gamut boundaries of the target medium.

\end{minipage}
}%
\end{figure*}

\section{HDR Gamut Management Framework}
\label{ScAlgorithm}

\begin{figure*}[t]
\centerline{
\includegraphics[width=\textwidth]{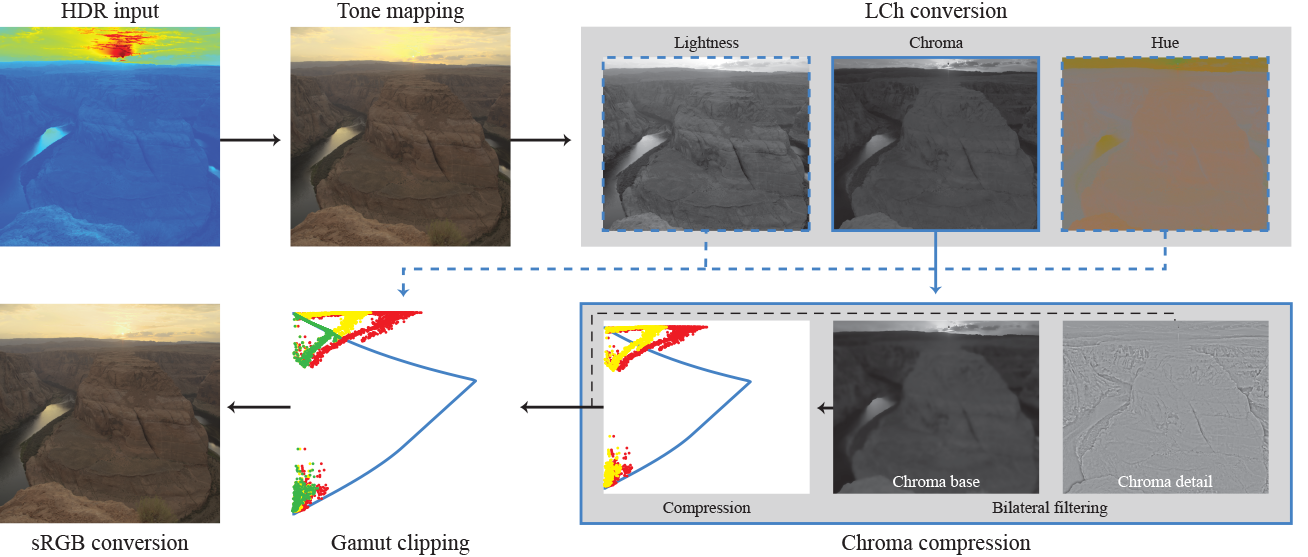}
}
\caption{Overview of our framework. The input HDR image (here illustrated with a
heat-map of luminance values) is first tone mapped and then converted to the
$LCh$ color space. Chroma values are then filtered using the bilateral filter
and the base layer is compressed. Finally, a gamut clipping step ensures that
the compressed chroma and lightness values fit within the target gamut while
minimizing appearance changes. (Red: input, Yellow: Chroma compressed, Green:
Final result)}
\label{fig:Algorithm}
\end{figure*}

Our framework incorporates tone mapping, chroma correction, and gamut management
to process an HDR image for a given output gamut. In this section, we will
discuss the components of our solution in the context of existing dynamic range
compression algorithms. The workflow is shown in Figure~\ref{fig:Algorithm}.

The input to our pipeline is an HDR image $I$ given in linear XYZ coordinates.
First, the luminance channel of the image, denoted $I(L)$, is compressed using
any existing tone mapping operator (TMO). The chroma channel $I(C)$ of the
resulting image is then corrected with our chroma compression algorithm to
correct for unwanted saturation due to the tone mapping process. Finally, image
chroma and luminance values $I(C,L)$ are processed with a gamut management step
to ensure that all pixels fit within the target gamut boundaries, denoted
$G(C,L)$, while minimizing appearance changes in the image.

\subsection{Luminance Compression}

As we want to ensure that our chroma compression and subsequent gamut
management steps correct any issues that the tone mapping process may
have caused, the first component of our framework is the luminance
compression. Our framework has been designed to easily integrate with
existing TMOs and in the following we describe the steps for
integrating the commonly used Photographic operator~\cite{Rein2002a}:
\begin{steps} 
\item The input HDR is first
converted to the color space expected by the TMO (in this case $Yxy$).
\item The tone mapping curve is applied on the luminance Y, obtaining the
compressed value $Y_{c}$. Both local and global algorithms can be applied at
this point as our framework poses no restrictions on the type of processing. 
\item The compressed luminance is inserted into the image and the result is
converted back to the $XYZ$ color space.
\item Finally, the tone mapped image $I$ is normalized, such that the maximum
$Y$ value is 100, to allow for further processing. In the case of the
Photographic operator, output luminance values are between 0 and 1, and as such
require scaling to the range expected by the color space used in further
processing. This is discussed further in the following section.
\end{steps}
 
Although this process allows for flexibility in the choice of
luminance compression, it comes at the cost of increased computational
complexity due to additional color space transforms, as existing TMOs
are not necessarily designed to operate in a color space that enables
gamut manipulations. As an alternative solution, we have designed a
compression solution that follows a similar scheme to our chroma
compression method, which is described in the side-bar ``Lightness
Compression using Cusp Alignment''.

\subsection{Gamut Boundary Computation}

Our algorithm relies on the idea that to avoid undesired shifts in
chroma and hue, as well as to avoid uncontrolled clipping for
out-of-gamut colors, one should work in a perceptually decorrelated
color space where these components are separated. A natural choice for
this is the CIE $L^*C^*h^*$ color space, which is the cylindrical
representation of CIE $L^*a^*b^*$ \footnote{In the remainder of this document, we
will refer to these spaces as LCh and LAB for brevity.}. These two
color spaces are commonly used in traditional gamut mapping
algorithms~\cite{Morovic2008}.

\begin{figure}[th]
\centering
\includegraphics[width=\columnwidth]{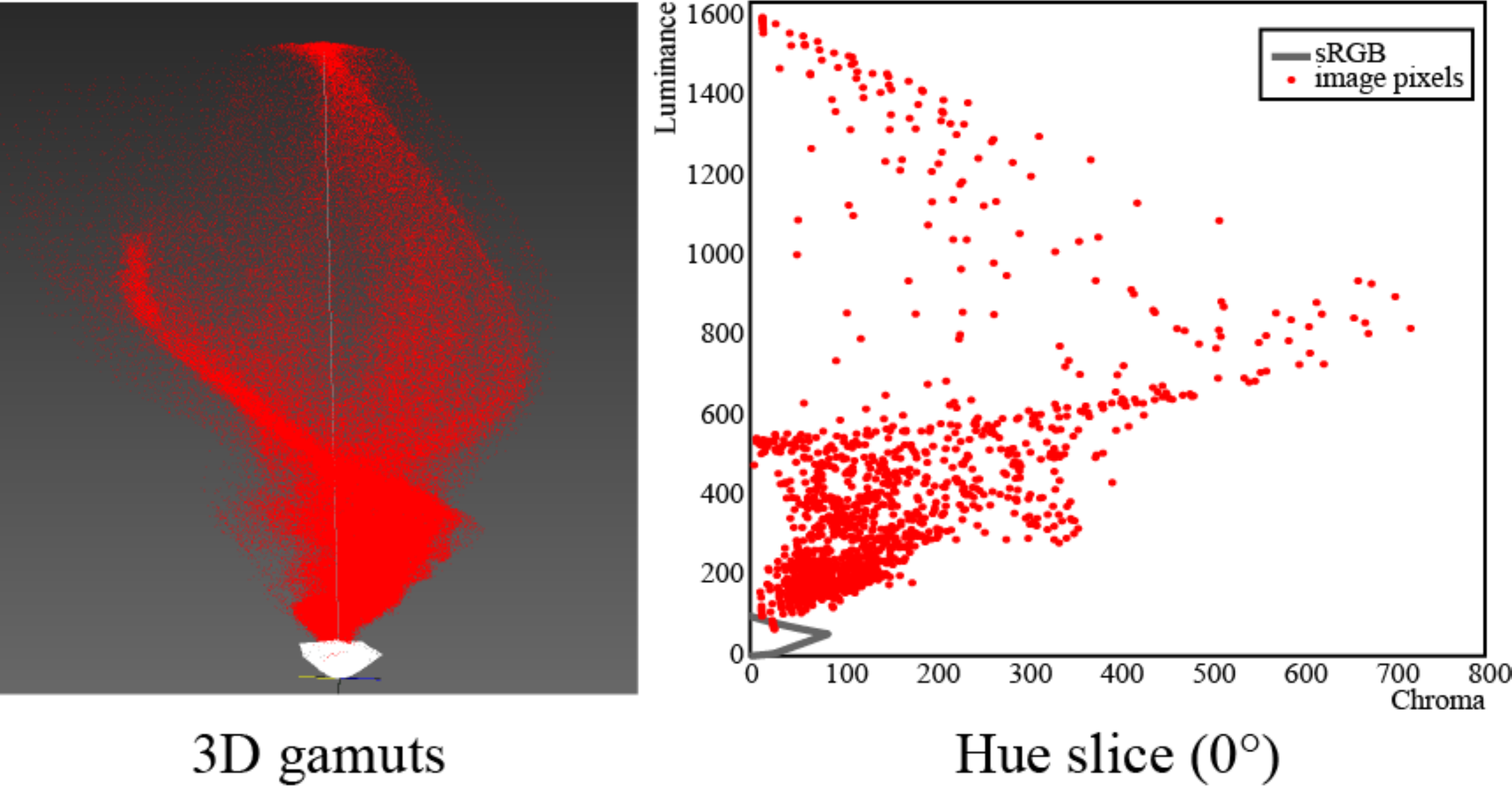}
\caption{Boundaries of the source gamut (input HDR image - red) and the target gamut (sRGB color space - white).}
\label{fig:Gamuts}
\end{figure}

To determine the correct compression amount and to assess whether a
given pixel can fit within the output gamut, we need to know the
boundaries of that gamut. In this paper, we assume that the target
gamut is sRGB and as such use a D65 white point when converting to
LAB. Our algorithm, however, can accommodate any alternative gamut and
corresponding white point. The source or input gamut is the set of colors of 
the input HDR image. Computing the boundaries of both gamuts we have followed the methodology described
in~\cite{Morovic2008} 
to obtain the boundaries of the sRGB gamut in
LCh coordinates, which take the form of a triangular cusp along the
chroma-lightness plane for each hue value.
Figure \ref{fig:Gamuts} shows the extreme differences between the source (red)
and the target (white) gamuts that may occur in HDR imaging.   
Between source and target, the differences in both lightness and chroma channels exceed a factor of 10.

\subsection{Chroma Compression}
\label{sec:Gamut Compression}

When considering HDR imagery, a large number of pixel values may be outside the
destination gamut in terms of chroma, generating unwanted hue shifts
if clipped in an uncontrolled manner. 

Additionally, it has been shown that tone compression along the luminance
dimension tends to create an over-saturated appearance in images. This is
illustrated in Figure~\ref{fig:Teaser}. To correct these issues, we are
proposing two methods that compress the chroma values $I(C)$ in the image in a
content-dependent manner.


\subsubsection{Hue-Specific Method}
\label{sec:Hue-Specific}

\begin{figure*}[th]
\centering
\includegraphics[width= \textwidth]{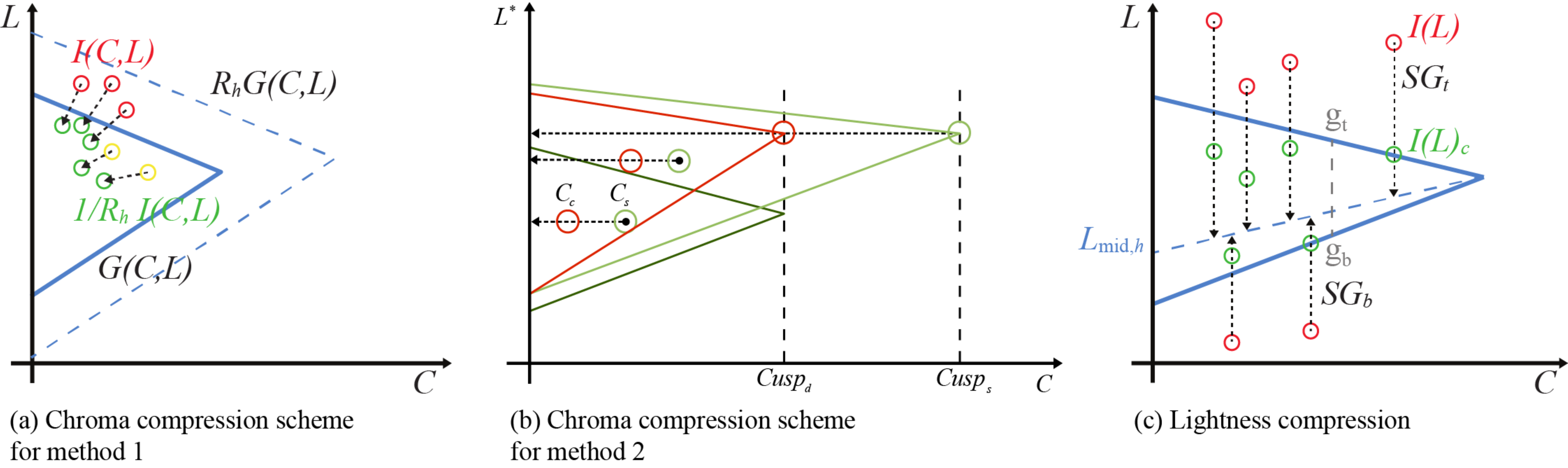}
\caption{Compression scheme presented in this paper for the (a) Chroma compression (Hue-Specific),
(b) Chroma compression (Global) and (c) Lightness compression.}
\label{fig:CLcompression}
\end{figure*}

This method compresses the chroma values $I(C)$ in the image using a two
step process.  The first step for our algorithm is to determine a
scaling factor $R_h$ for each hue value $h = \left[1^\circ \ldots
360^\circ\right]$, leading to a vector $\textbf{R}$. To achieve that,
we scale the gamut boundaries $G_h(C,L)$ until they enclose all pixels
within that hue slice. Formally, we initialize the scaling factor for
a hue slice $R_{h,0} = 1$. If there are pixels that are out of gamut
for that hue slice, at each iteration step $i$, we increment the scale
factor $R_{h,i}$ and scale the gamut boundaries as follows:
\begin{align}
R_{h,i}        & = R_{h,i-1} + d\\
G_{h,i}(C,L)    & = \left[
\begin{array}{cc}
R_{h,i} & 0\\
0 & R_{h,i}
\end{array}\right]G_{h}(C,L),
\label{eqScaleG}
\end{align}
where the increment $d$ is set to a small value (in all the results in
the paper $d = 0.1$). This process is illustrated in
Figure~\ref{fig:CLcompression}a.

While this scaling factor could be applied to the image values
directly to obtain a within-gamut result, in practice, using the full
chroma range of the source gamut would likely result in extreme
compression due to a few outlying pixels with extremely high
chroma. This is a common problem in luminance compression, where some
extremely bright highlights may lead to an over-compressed result and
is usually countered by compressing according to a percentile of the
range of values. In the case of chroma, if the compression takes into
account such pixels, the resulting image may be too desaturated.

To avoid this undesirable effect, we use a percentile of the chroma
range when computing $\textbf{R}$. We have found that the
percentile value required is content dependent---if the pixels that
require clipping are spread over the image, then a more aggressive
percentile value may be selected. If, however, the out-of-gamut pixels
are concentrated in a small number of regions, a more gentle approach
is necessary to ensure that no artifacts are created. In practice, we
determine the spread of out-of-gamut pixels by computing the number of
connected regions that they belong to and comparing them to the number
of pixels contained within them. We have found that a ratio less than
0.01 suggests that the out-of-gamut pixels are within a few connected
regions and therefore further clipping may lead to artifacts.

\begin{figure}[th]
\centering
\includegraphics[width=\columnwidth]{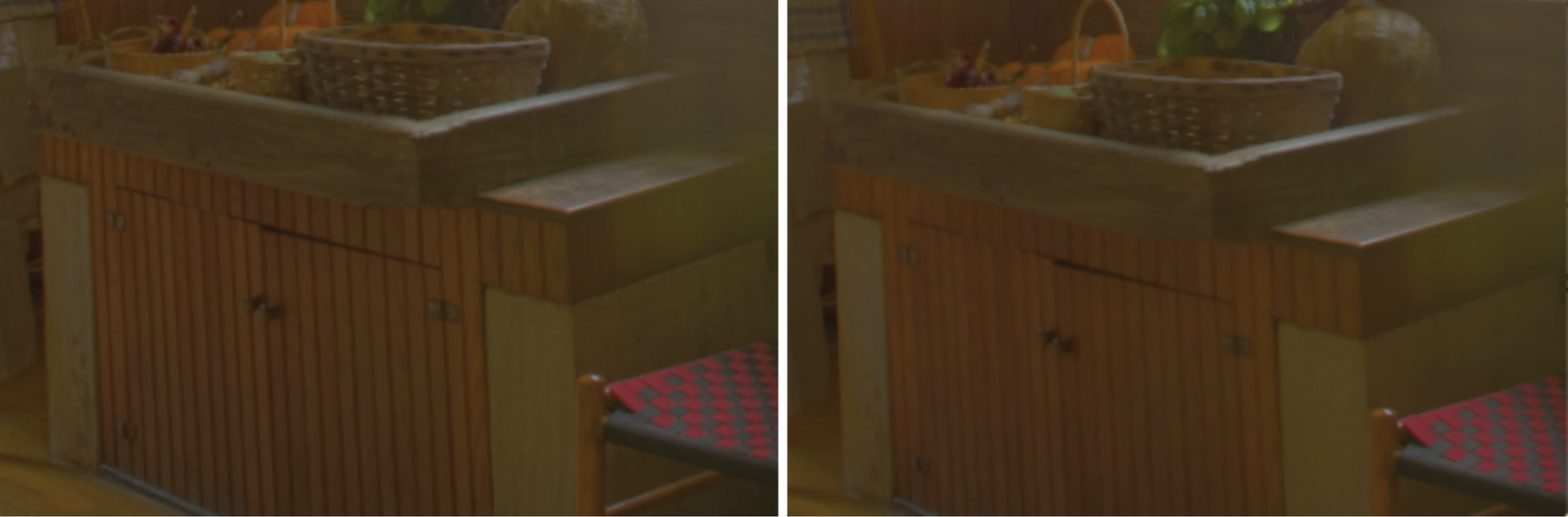}
  \caption{Results (left) with and (right) without bilateral filtering decomposition.}
\label{fig:BilateralC}
\end{figure}

To be able to keep fine details without smoothing edges, the chroma 
channel of the image $I(C)$ is first processed
with the bilateral filter ($\sigma_s=0.2\max(I_{width},I_{height}),\sigma_r = 0.05
\max(I(C))$) obtaining a base layer $I(C)_{\textrm{base}}$, which
is used in the above computation.
Either division or subtraction  can be used to separate the base and 
detail layers. We have verified that both methods are leading to similar 
results so as we have decided to use the division method for producing the detail layer
$I(C)_{\textrm{detail}} =I(C)/I(C)_{\textrm{base}}$. Figure \ref{fig:BilateralC} (left) shows how fine details 
are preserved, i.e. doors of the drawer and borders of the sink, when
bilateral filtering is applied to the chroma channel.

Even with these measures, small variations in content between adjacent
hue slices may lead occasionally to discontinuities in the final image
if $R_h$ is applied directly to each slice. Smoothing the scaling
vector $\textbf{R}$, to $\textbf{R'}$, will eliminate these
discontinuities.  To achieve this, one can use different type of
smoothing functions.  We have used four different smoothing functions:
\emph{lbox} (averaging box),  \emph{loess} (locally weighted
regression), \emph{rloess} (robust locally
weighted regression) and \emph{sgolay}
(Savitzky– Golay).  We find that the four smoothing functions produce 
results of similar quality, so the simpler and more
computationally efficient function can be used in our framework
(\emph{lbox}).  During the smoothing step, the circular nature of hues
is taken into account. This avoids the creation of boundaries between
the hue angles of 359 and 0.


Finally, image chroma within each hue slice $I(C)_{\textrm{base},h}$
is scaled as:
\begin{equation}
I(C)'_{\textrm{base},h} = \frac{1}{R'_h} I(C)_{\textrm{base},h}
\end{equation}
and the detail is re-injected to obtain the chroma-compressed image
$I(C)' = I(C)'_{\textrm{base}}\times I(C)_{\textrm{detail}}$, where
$\times$ indicates an element-wise multiplication. Note that the
reciprocal $\textbf{R'}$ needs to be used for the final compression as
$\textbf{R}$ was initially computed to expand the gamut boundaries
until they enclosed all pixels

\begin{figure}[th]
\centering
 \includegraphics[width=\columnwidth]{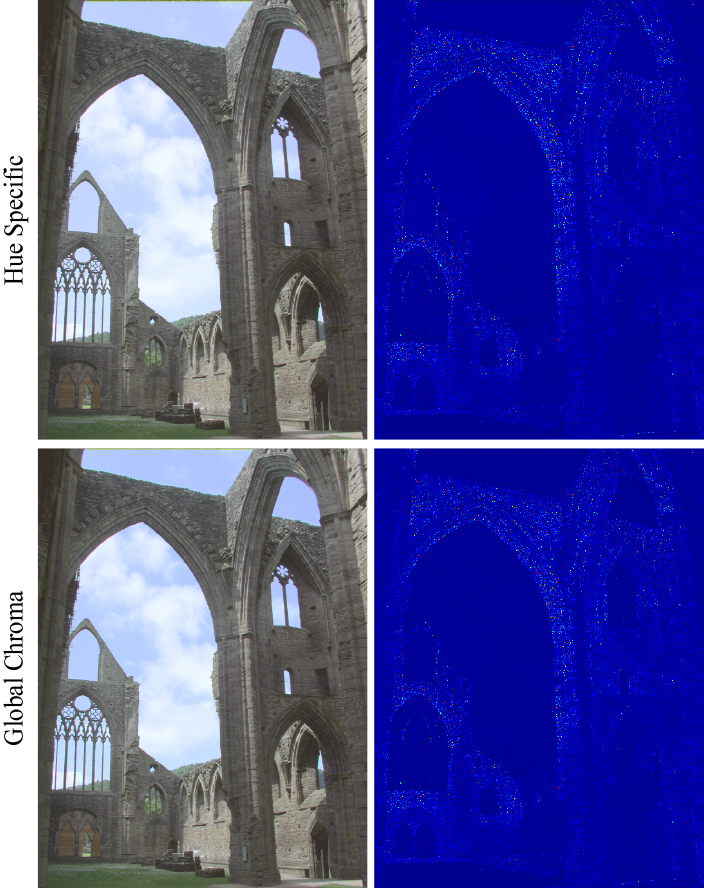}
  \caption{Results using the two chroma compression methods proposed in this paper and their hue differences maps.}
\label{fig:Chroma_Comp}
\end{figure}

\subsubsection{Global Method}

The hue-specific method described in the previous section can
successfully compress chroma, therefore
maximizing the use of the available gamut. This however comes at the
cost of increased computational complexity. At the same time, we may
observe that the dynamic range of the chroma channel is not extremely
high, when compared with the dynamic range of the display gamut. This
suggests that a linear compression scheme may produce good results
as shown in the hue difference maps of Figure~\ref{fig:Chroma_Comp}.
As such, we propose an alternative but simpler method based on similar
premises. We envisage that the hue-specific method would be suitable
for a post-production pipeline where accuracy is the primary goal,
while the linear method described in this section would be better
suited on the display-side of an imaging pipeline, where computational
resources are limited.

The global method is illustrated in
Figure~\ref{fig:CLcompression}(b). In this figure, the light and dark
green triangles represent the source and destination gamuts
respectively for a fixed hue angle. Our goal is to align the two
chroma cusps while maintaining lightness and hue. In the figure, this
would produce the red triangle which represents the compressed gamut.

Chroma compression is still applied on each hue angle from $0^{\circ}$
to $359^{\circ}$. Instead of using different amounts of chroma
compression for each hue however, which requires additional
computations, we compute the minimum of the cusps ratios across all
hues. This value corresponds to the maximum necessary compression.

Similar to the hue-specific method, chroma compression is applied
to the base layer of the chroma channel. We compress the source chroma
$I(C)_{\textrm{base},h}$ to yield the compressed
chroma $I(C)'_{\textrm{base},h}$ as follows:
\begin{equation}
I(C)'_{\textrm{base},h}  = I(C)_{\textrm{base},h} \min_{h\in [0^\circ,359^\circ]} \left[\frac{\textrm{Cusp}_{d,h}}{\textrm{Cusp}_{s,h}}\right].
\label{eq:C- Compression}
\end{equation}
where it is noted that this is only a compression when $\min
\left[\textrm{Cusp}_{d,h}/\textrm{Cusp}_{s,h}\right] < 1$.

We may face similar issues as in the hue-specific method when using
the full chroma range of the source gamut. This will produce extremely
compressed chroma results in some cases due to the use of a few outlying
pixels with extremely high chroma values. To avoid this problem, we
have adopted the same percentile approach used for the first method. 
This solution avoids over-compression but requires an
additional step for managing the few pixels that may remain outside
the gamut boundary, which is discussed in the next section.

\subsection{Gamut Clipping}
\label{sec:Merging}

\begin{figure}[th]
\centering
  \includegraphics[width=\columnwidth]{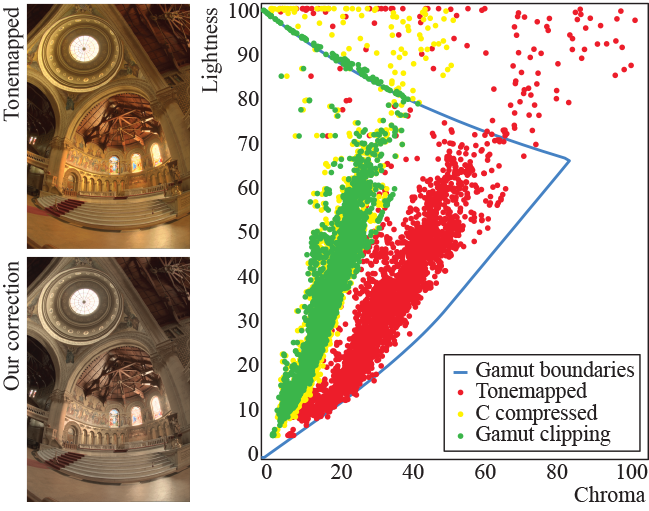}
  \caption{Here we plot the chroma values (x-axis) of pixels within a hue
  slice ($h=60^\circ$) against their lightness values
  (y-axis). After tone mapping the lightness channel of the image,
  many pixels (shown in red) are still outside the gamut
  boundaries. Scaling the chroma channel moves pixels closer to the
  gamut boundaries (yellow), while the last step in our framework
  ensures that pixels are mapped to values within the target gamut
  (green). The resulting image after tone mapping is shown at the top
  left and after our correction and clipping step at the bottom
  left.}
\label{fig:ClipPlot}
\end{figure}

Since chroma and lightness values are so far processed independently,
we cannot guarantee that all pixels will be within the target gamut
boundaries. This is shown in Figure~\ref{fig:ClipPlot}, during tone mapping 
pixels are compressed through the lightness direction. However, pixels are 
guarantee to have maximum lightness values equal to 100 nits, 
but may still be outside of the target gamut boundaries (red pixels). Applying chroma
compression does not solve this problem (yellow pixels).
To have all pixels within the target gamut boundaries, without further modifying pixels already
in-gamut, a clipping step is employed. As image pixels may be out of
gamut both in terms of chroma and lightness, as shown in
Figure~\ref{fig:ClipPlot}, pixel values need to be clipped along both
dimensions creating a trade-off between changes in lightness or in
chroma for each pixel.

Although many proposals exist for defining the clipping line along
which pixels should move, they are designed for scenarios where the
input and target gamuts differ in terms of chromatic primaries used
rather than luminance range. In our specific case however, we have
found that most out-of-gamut pixels tend to be bright, highly
chromatic pixels (red pixels outside the gamut boundaries in
Figure~\ref{fig:ClipPlot}).

\begin{figure}[th]
\centering
\includegraphics[width=\columnwidth]{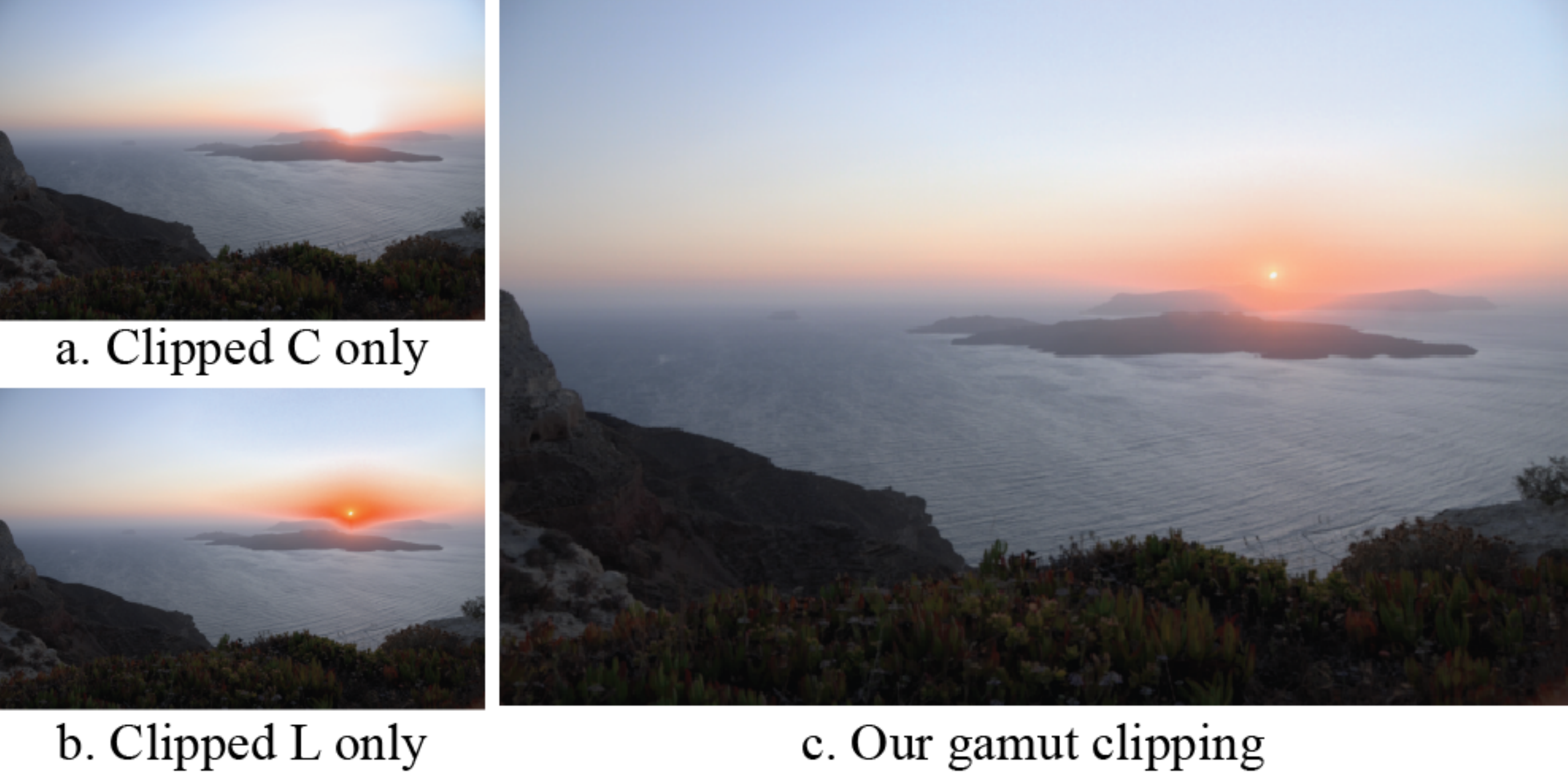}
\caption{Three different clipping solutions were used to process the same image.
Since most of the out-of-gamut pixels reside in the top right corner of the cusp,
clipping along only the chroma direction (a) severely desaturates these pixels.
In contrast, clipping only along the lightness direction (b) leads to an
unnatural, oversaturated appearance. The result of our interpolated solution is
shown in (c), where the image has maintained a natural appearance while all
pixels fit within the gamut boundaries.}
\label{fig:CclipVSLclip}
\end{figure}

Because of the narrowing of the cusp near high-L values, a delicate
balance between chroma and lightness adjustments is necessary to
ensure that the resulting image appearance does not change. This is
demonstrated in Figure~\ref{fig:CclipVSLclip}: the area around the sun
either desaturates completely as the lightness values are near the
peak of the cusp (Figure~\ref{fig:CclipVSLclip}a) or it looks too saturated
because its lightness is decreased with no corresponding chroma
changes (Figure~\ref{fig:CclipVSLclip}b).

Instead, we propose a middle-ground between these two extremes. In a
given hue slice, for a pixel $p \in I'$ we determine a point along the
gamut boundaries $p_{\textrm{clipC}}$ such that $p_{\textrm{clipC}}(C)
\in G(C)$ and $p_{\textrm{clipC}}(L) = p(L)$. Similarly, a value
$p_{\textrm{clipL}}$ is determined, where $p_{\textrm{clipL}}(C)$
remains unchanged and $p(L)_{\textrm{clipL}}$ moves to the gamut
boundary. Once these two points have been computed, linear
interpolation is performed to map the out of gamut pixel to the
corresponding gamut boundary of the destination gamut.

\begin{figure*}
\fcolorbox{black}{lccol}{
\begin{minipage}[t]{\textwidth}
\section*{Side-bar: Lightness Compression using Cusp Alignment}
\label{sec:Lcompression}

Although the main goal of this paper is to present our framework for
managing the gamut mismatches that tone mapping causes in terms of the
resulting image chroma, we have found that our approach can be
directly extended to compress the lightness channel, leading to an
integrated luminance and color gamut management framework and
minimizing the number of color space conversions necessary.
To compress the lightness channel $I(L)$, we process each hue slice
separately similar to the process described in Section~\ref{sec:Gamut
Compression}. The compression scheme is depicted in
Figure~\ref{fig:CLcompression}(c).

Specifically, we follow these steps for each hue slice:
\begin{steps}
\item We first find the global parameters that express the maximum vertical (lightness)
distance from the destination gamut. The distance at the top is named $SG_{t}$
and at the bottom $SG_{b}$ (see Figure~\ref{fig:CLcompression} (c)). Both
values are set to $0$ when all pixels of the source gamut are already inside the
destination gamut.

\item The ``middle'' line $L_{\textrm{mid}}$ is computed for the cusp of each hue
slice by:
\begin{equation}
L_{\textrm{mid}} = g_{b} + (g_{t} - g_{b})\frac{SG_b}{(SG_{t}+SG_{b})},
\label{eq:Middle Line}
\end{equation}
where $g_{b}$ and $g_{t}$ are the bottom and the top values of the
destination gamut. Note that this equation shifts the middle line
toward $g_b$ for large $SG_t$, effectively compressing more of the
image and toward $g_t$ for large $SG_b$, in which case more pixels are
scaled linearly. That is, it adaptively determines a threshold that
separates the source gamut into two regions that can be thought of as
``light'' and ``dark'', with a magnitude for each determined by the
ratio $SG_t$:$SG_b$. We treat each of these regions separately. 

\item In the ``light'' region, i.e. for points above the $L_{\textrm{mid}}$,
the lightness is compressed as follows:
\begin{align}
I(L)_c &= a_t + b_t*F(I(L)-L_{\textrm{mid}}).
\label{eq:LT-Compression}
\end{align}
The compression factors $a_t$ and $b_t$ are computed by $a_t = L_{\textrm{mid}}$ and $b_t = \frac{(1-w)(g_t - L_{\textrm{mid}}) + w(100 - L_{\textrm{mid}})}{N}$, respectively, where the normalization factor is $N   = F(g_t + SG_{t} - L_{\textrm{mid}})$  and the weight $w$ is computed as $w   = \frac{I(C)}{I(C) + \max(G(C))}$.

Here $F(x)$ represents a non-linear compression function, which is
applied to pixels above $L_{\textrm{mid}}$. This could be any desired
tone curve, including for instance sigmoidal compression or basic
compressive functions such as roots and logarithms. Note that
Equation~\ref{eq:LT-Compression} compresses the range
$[L_{\textrm{mid}},g_t + SG_t]$ to $[L_{\textrm{mid}},
b_tN+L_{\textrm{mid}}]$.  Consequently, pixels with really high chroma
and lightness may be mapped above the gamut boundary, and therefore
clipped in the following stage of our framework. Since the clipping
process takes into account both $C$ and $L$ values, this ensures that
such pixels will still remain bright in the final image.

\item For points below $L_{\textrm{mid}}$, linear compression is used:
\begin{equation}
I(L)_c = a_b + b_b*(I(L)-L_{\textrm{mid}}).
\label{eq:LB-Compression}
\end{equation}
In this case, the parameters are computed as $a_b = g_b$ and $b_b = \frac{L_{\textrm{mid}}-g_b}{L_{\textrm{mid}}}$.

Note that here we compress the range $[g_b - SG_b,L_{\textrm{mid}}]$ to $[g_b,L_{\textrm{mid}}]$.  Our method effectively corresponds to a non-linear compression with a linear ramp in dark areas. Such a behavior is commonly used in  film. In our case however, the gamut of the image is explicitly considered to guide this compression scheme.
\end{steps}

\vspace{8pt}
To improve  visibility in dark regions, we adopt a similar technique used for
the chroma compression by making use of a percentile and processing $I(L)$ with
the bilateral filter before compression. An example of our tone curve for the
base layer compared with other global
TMOs  is shown in
Figure~\ref{fig:ToneCurve}~\cite{Drago03a,Ashikhmin2002,Rein2002a}.

\vspace{8pt}
\centerline{
\begin{minipage}[H]{\textwidth}
\centering
\includegraphics[width=0.9\textwidth]{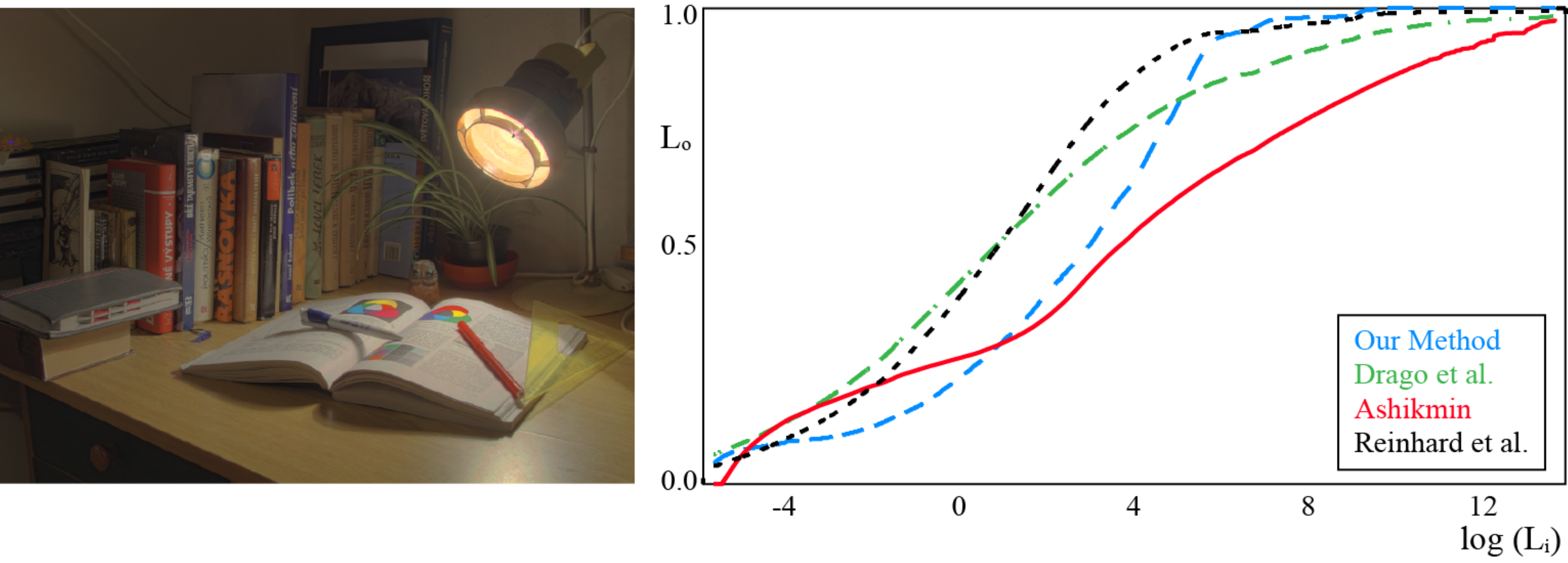}
\captionof{figure}{Tone curves making use of (blue) our method (HDR Gamut) and
  three global TMOs - (red) Ashikmin~\cite{Ashikhmin2002} -
  (green) Drago et al~\cite{Drago03a} - (black) Reinhard et
  al.~\cite{Rein2002a}.}
\label{fig:ToneCurve}
\end{minipage}
}

\end{minipage}
}%
\end{figure*}


\begin{figure*}[t]
\centering
\includegraphics[width=\textwidth]{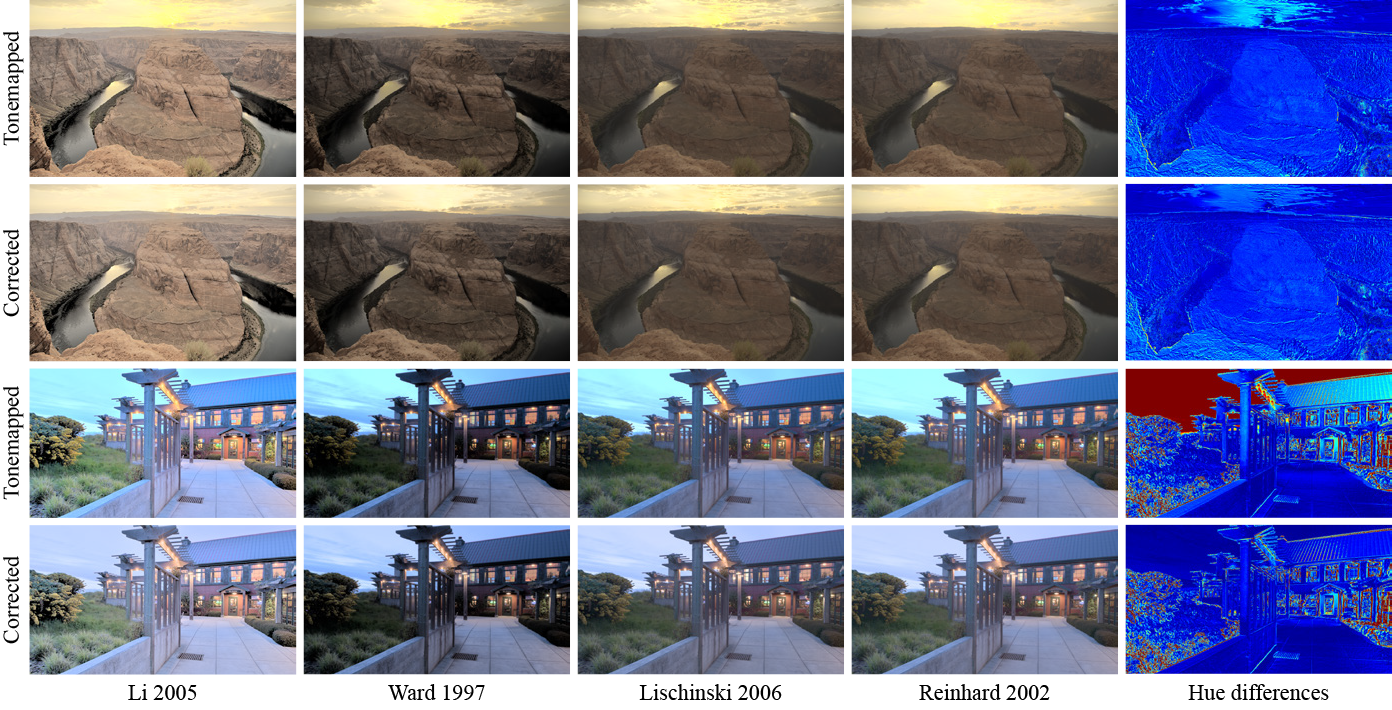}
\caption{Shown here are HDR images that were processed with the TMOs by Li et
al., Ward et al., Lischinski et al., and Reinhard et al. The output of the TMOs with
no further processing is shown at the top, while the bottom row shows the output of the TMOs integrated with our chroma compression technique. The rightmost column shows hue
differences between the TMO output and the HDR original (top) and between our corrected result and the HDR original (bottom) for the TMO by Reinhard et al. Hue differences are reduced for other TMOs as well.}
\label{fig:TMOIntegration}
\end{figure*}

\section{Results}
\label{sec:Results}

We have validated our technique using several challenging HDR images,
demonstrating its benefits over existing techniques, including gamut
mapping solutions, color correction methods, and color appearance
models (CAMs). Additionally, we show the flexibility of our approach
to be used with existing TMOs, without degradation of the details or
introducing unwanted artifacts. We evaluate the quality of our
reproduction by measuring hue changes as well as through a
psychophysical study comparing our chroma compression method with
alternative techniques.  All the results shown in this section assume
sRGB primaries for both input and output. They have been gamma
corrected using the sRGB gamma correction equation, and use the chroma
compression method specified in Section~\ref{sec:Hue-Specific}.

\begin{figure*}[t]
\centering
\includegraphics[width=\textwidth]{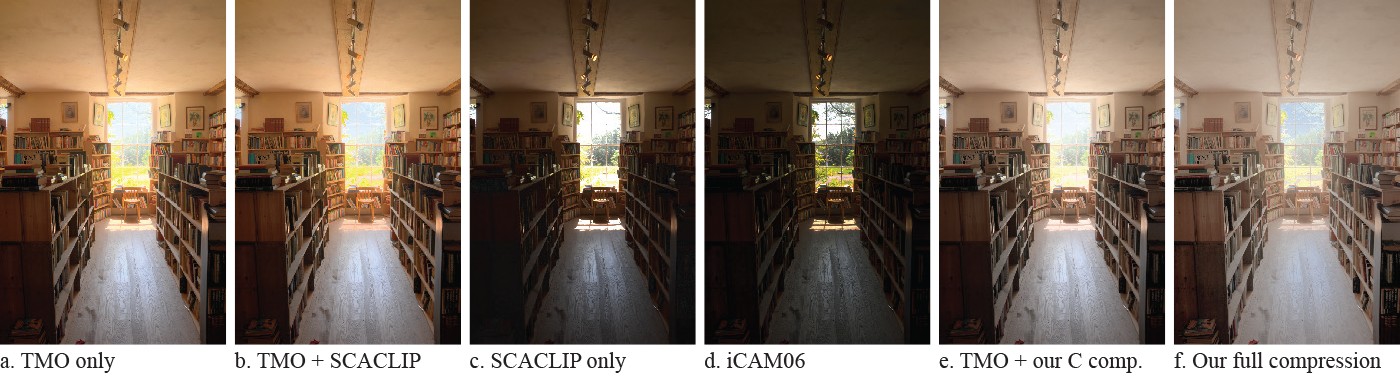}
\caption{An HDR image is shown after being processed with different compression solutions. 
(a) Tone mapped with the Photographic operator~\cite{Rein2002a} 
(b) Tone mapped and then processed with the SCACLIP gamut mapping 
method~\cite{Bo+07} (c) Processed only with SCACLIP (d) Processed
 with iCAM06~\cite{Kuang2007} (e) Tone mapped and processed with our
 chroma correction step (f) Processed with our integrated lightness and chroma 
compression.}
\label{fig:ComparisonBookstore}
\end{figure*}

\begin{figure}[t]
\centering{
\includegraphics[width=\columnwidth]{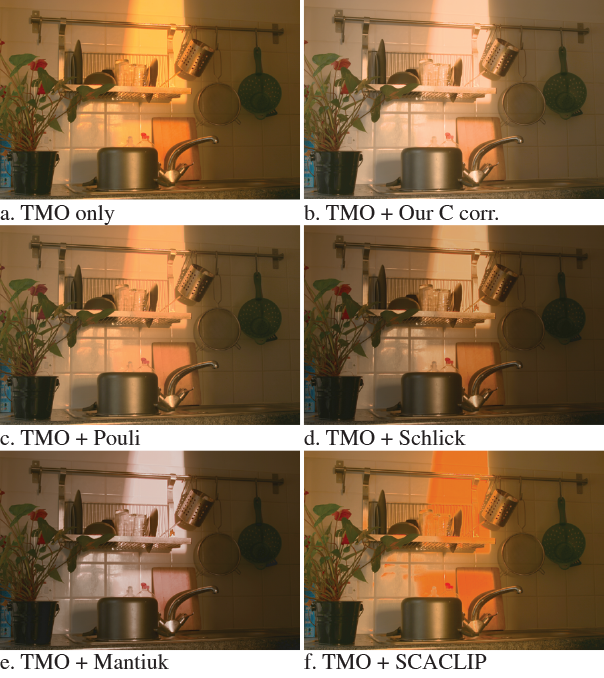}
}
\caption{The tone mapped image (a) was corrected using our method (b) as well as existing color correction solutions (c,d and e)~\cite{Pouli:2013,Sc94,Ma+09}.These methods aim to
correct the over-saturated appearance resulting from many tone mapping solutions and operate as a post-process on the image, without any gamut considerations. The result of combining the SCACLIP gamut mapping method with a TMO is shown in (f).}
\label{figColorCorrections}
\end{figure}

Our approach can handle challenging images, producing natural results that
preserve details in the image while fitting within the target output gamut.
Figure~\ref{fig:Teaser} shows the result of processing such an image 
with our framework as well as with other techniques (see supplemental material 
for more images). Existing tone mapping techniques (in this case the Photographic
operator~\cite{Rein2002a}) can effectively compress the luminance in the image
but lead to an oversaturated appearance, with many pixels still out-of-gamut 
(e.g., colors of the macbeth color checker).

\begin{figure}[t]
\centering
\includegraphics[width=0.5\textwidth]{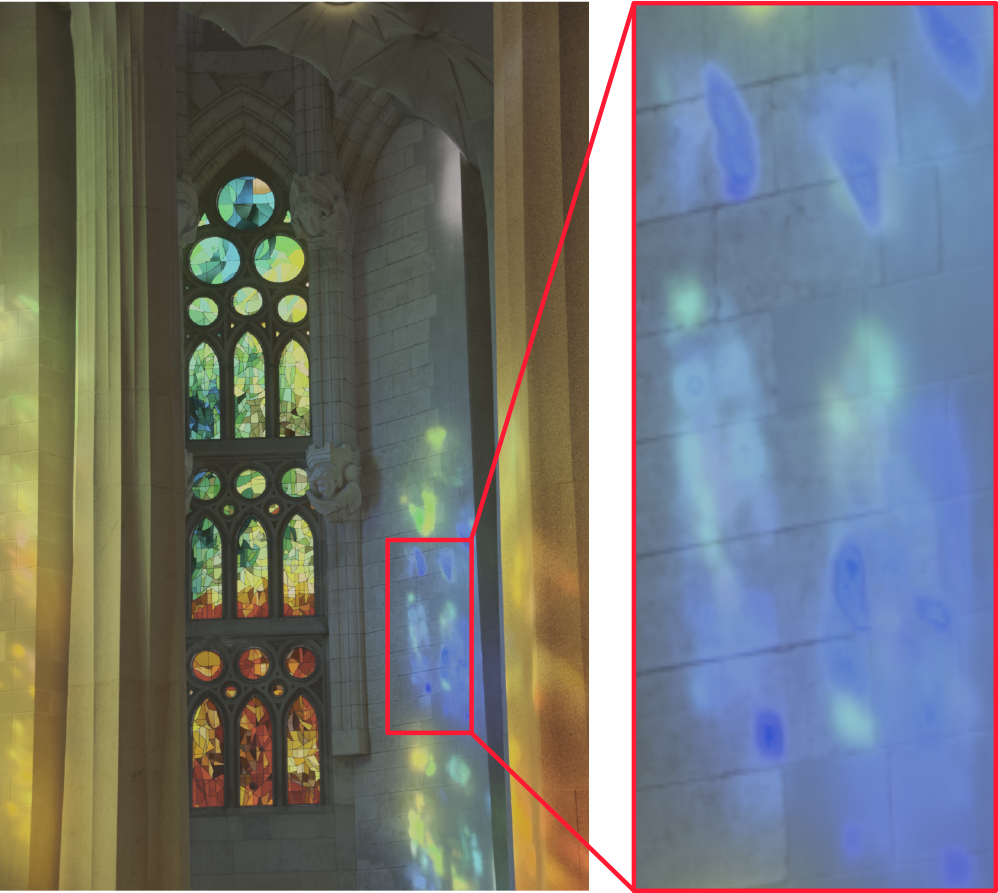}
\caption{An example showing that the SCACLIP gamut mapping method~\cite{Bo+07},
 integrated with an existing TMO~\cite{Rein2002a} may introduce artifacts.}
\label{fig:SCACLIPArtefacts}
\end{figure}

Although a gamut mapping solution such as SCACLIP~\cite{Bo+07} can control this
issue by moving pixels within the gamut boundaries, it may amplify the appearance of over-saturation (Figure~\ref{fig:Teaser}e) or it introduces artifacts as shown in §Figure~\ref{fig:SCACLIPArtefacts}. At the same time,
although such a gamut mapping approach modifies lightness values, it cannot
sufficiently compress the extreme dynamic range of this image if used alone as
seen in Figure~\ref{fig:ComparisonBookstore}c.

In contrast, our framework combines the advantages of tone mapping and gamut
mapping (Figure~\ref{fig:Teaser}b and Figure~\ref{fig:ComparisonBookstore}e). Our solution also allows for
flexibility in the choice of compressive function: different functions can lead
to different image appearances as shown in Figure~\ref{fig:TMOIntegration}. Despite
the very different tone mapping styles, our chroma correction leads to
a consistent treatment of colors in the images.

Finally, Figure~\ref{figColorCorrections} shows a comparison between our chroma correction 
and other color correction solutions that are typically applied as a post-process to tone
 compression as well as the result of SCACLIP after tonemapping. Note that the three 
color correction methods shown do not consider the gamut boundaries and therefore 
may lead to out-of-gamut pixels (in our experiments, we found that this often includes 
more than $10\%$ of the image pixels).

\subsection{Hue differences}
\label{sec:hue}

Ideally, compressing separately lightness and chroma at fixed hue should not
affect hues. To assess whether our algorithm achieves that, we have
evaluated our results using color difference metrics. Typically, color
differences are computed using $\Delta E$ color difference metrics, which take
into account both luminance and chromatic differences.
In our case, a metric capable of separating
luminance, chroma, and hue is necessary as we are only interested in preserving
the hue while luminance and chroma are being compressed.

Although color difference measurements are commonly performed in the
LAB color space, it is known that LAB is not
hue-linear across all hues. Instead, we use an optimized $I'P'T'$ space for color difference comparisons \cite{Shen:2009}. 
This space is scaled and rotated with respect to $IPT$ such that color
differences in this new space are directly comparable with other color
difference metrics, while preserving hue linearity.

In $I'P'T'$, a cylindrical space is then computed, where lightness
$\Delta I'$ and hue $\Delta h$ differences can be calculated. We use
$\Delta h$ instead of the perceptually scaled CIE $\Delta H$ metric,
as the latter scales hue differences by colorfulness to account for
perceptual effects, and is therefore not suitable for our particular
application.

As hue is defined on a circle, we compute $\Delta h$ for a given pair
of hues $h_t$ and $h_c$ as follows:
\begin{equation}
  \Delta h = \min(|h_t - h_c|, |\min(h_t,h_c) + 2\pi -
  \max(h_t,h_c)|)
\end{equation}

\begin{figure}[t]
\centerline{
\includegraphics[width=\columnwidth]{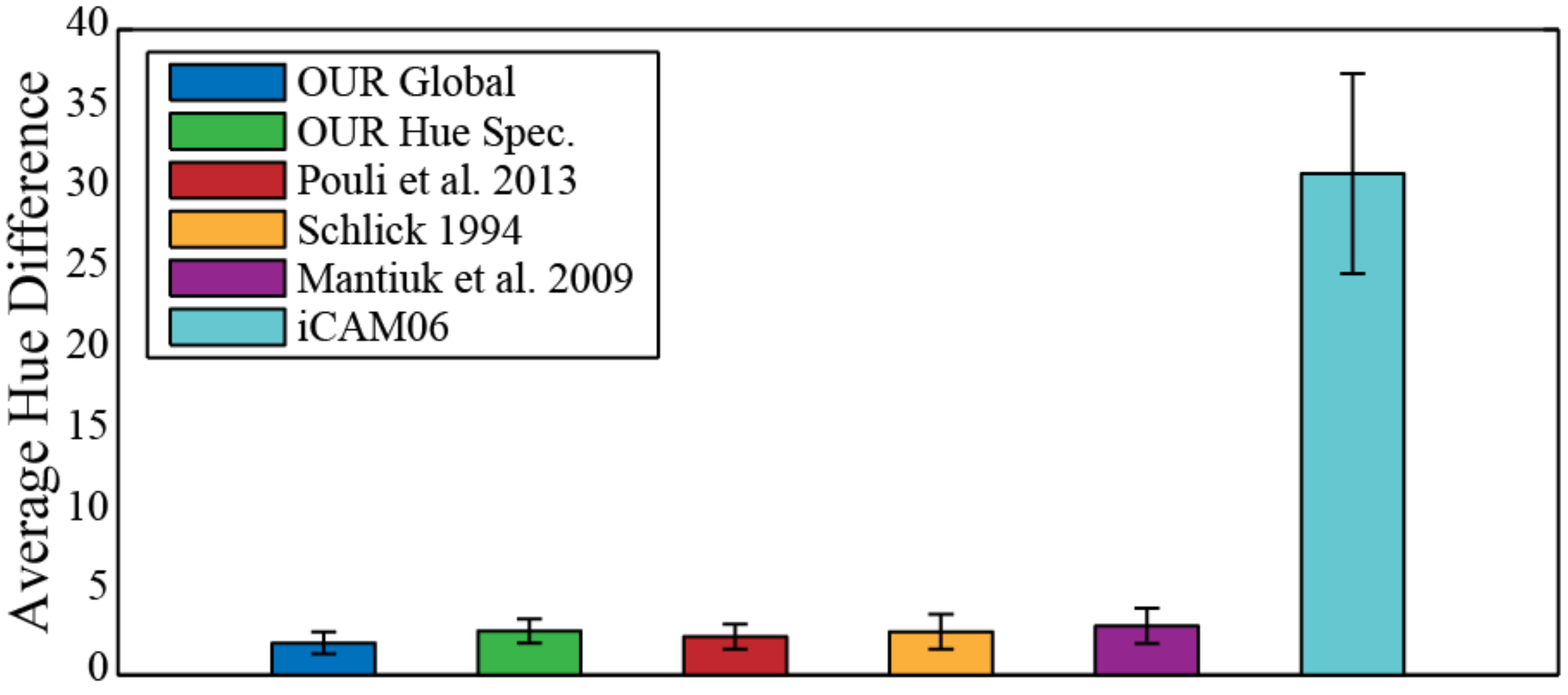}
}
\caption{Hue differences and standard error over 20 images for
    the two proposed methods as well as alternative techniques.}
\label{fig: DeltaE}
\end{figure}

Figure \ref{fig: DeltaE} shows the graph of $\Delta h$ computed over 20
images for the following methods \cite{Pouli:2013,Ma+09,Sc94}
~\cite{Kuang2007}. The methods have been used to adjust saturation
after the luminance dynamic range has been adjusted to the display
capability using the photographic operator \cite{Rein2002a}. When using the iCAM06 method~\cite{Kuang2007},
the luminance dynamic range is compressed with its own compression
technique.
While iCAM06 was developed to reproduce the correct appearance of colors
under different illumination conditions and in the context of HDR imaging, it is introducing 
a large hue shift when compared with the proposed technique. We note that our method does 
not introduce such hue shifts.

\subsection{Phsychopysical Evaluation}
\label{sec:Experiment}

Typically, when compressing the gamut of an image for a particular
display, the goal is to preserve the color appearance and general
quality of the image as much as possible, while conforming to
a more limited gamut. To assess the ability of our method to
preserve image quality despite gamut restrictions, we have
performed a psychophysical study. We used a two-alternative
forced choice design, with the linearly scaled HDR reference shown at the same time, allowing us to assess the fidelity of the color reproduction of the processed images compared to the HDR input. 

There were 13 participants (8 males and 5 females) in this experiment, who were between 22 and 25 years old, and all had normal or corrected-to-normal vision as well as normal color vision. Based on a pilot study comparing our chroma compression method against
the iCAM06 model~\cite{Kuang2007}, the SCACLIP gamut compression
method~\cite{Bo+07}, and the correction methods of Mantiuk et al.~\cite{Ma+09} and Pouli et al.~\cite{Pouli:2013}, we opted for comparing our method against the method of Mantiuk et al.~\cite{Ma+09} and the method of Pouli et al.~\cite{Pouli:2013} in a complete experiment, as the remaining methods were significantly less preferred. We also included the uncorrected tone mapped image in our evaluation which serves as our baseline result. 

In our evaluation, the 13 participants viewed the differently
processed results for 6 different scenes in pairwise comparisons
between the alternative methods and were asked to select in each
case the image that has the reproduction of color closest to the
HDR image, which was shown as linearly scaled on the same screen. A colorimetric calibrated NEC MultiSync P241W sRGB monitor was used for this experiment. 
As it is not possible to accurately reproduce the HDR ground truth on this monitor, users could control the exposure for the HDR image manually, allowing them to more accurately compare the processed results with the ground truth image. 

\begin{figure}[t]
\centering
\includegraphics[width=\columnwidth]{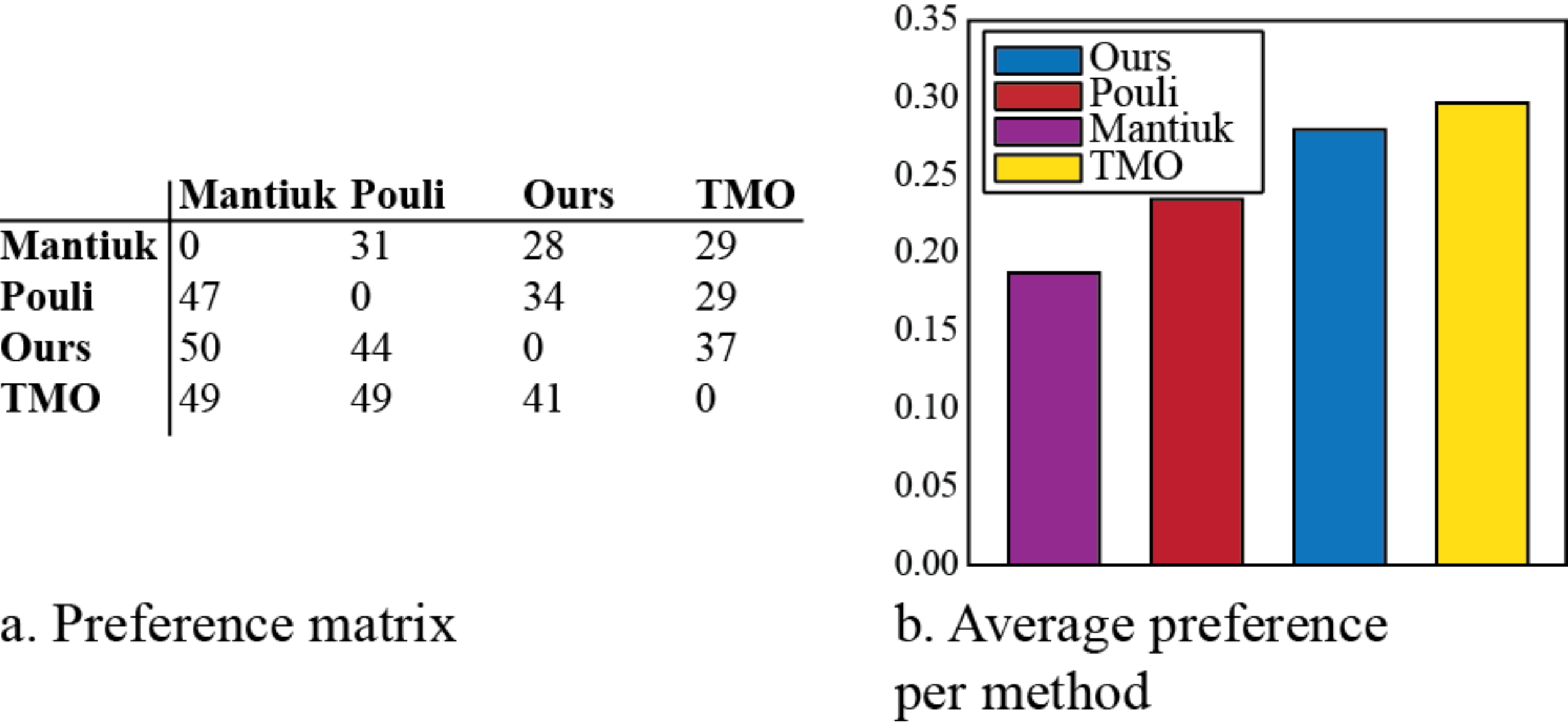}
\caption{(a) Aggregated preference results from our experiment for the different tested methods. Each matrix cell represents the number of times the method in each row was chosen over the method in each column. (b) Average total preference per method.}
\label{fig:expResults}
\end{figure}

Detailed results are shown in Figure~\ref{fig:expResults}. We performed significance analysis on the experiment results computing the $\chi^2$ value and agreement coefficient. Overall, the $\chi^2$ value was 21.23 with an agreement coefficient of 0.033. At a 0.05\% significance level, the critical $\chi^2$ value is 12.59, indicating that our results are significant. We note however that the agreement between participants was not very high. By further analyzing our results for individual images, we observed that participants agreed in their choices for some images, while agreement was lower for other images. We also observed that images with higher agreement were generally more saturated and colorful. Based on that, we repeated our analysis, but splitting the images into two groups depending on overall saturation: for the saturated group, $\chi^2=53.38$ (agreement coeff. 0.315) while for the less saturated group, $\chi^2=6.69$ (agreement coeff. 0.002). These results suggest that the benefit of our method is more visible in more colorful images with higher saturation, which is to be expected since these images are more likely to have out-of-gamut pixels. 

Overall, we found that although the method of Mantiuk et al.~\cite{Ma+09} was chosen significantly fewer times, all other alternatives (our method, method of Pouli et al. \cite{Pouli:2013} and uncorrected tone mapped only) were not found to be significantly different, suggesting that our method was found similarly visually pleasing by our participants as the uncorrected tonemapped results, while allowing for a controlled management of the gamut. As mapping out-of-gamut pixels inside the available gamut always presents a trade-off in visual quality, our method could be expected to offer a somewhat lower visual quality compared with the method of Pouli et al.~\cite{Pouli:2013}. However, our results show this not to be the case. The advantage of the present method relative to \cite{Pouli:2013} is therefore the inclusion of unobtrusive gamut management.

Given that both the image and the display color space in this case was sRGB, gamut management was only necessary due to potential out-of-gamut issues introduced by the tone mapping process. The necessity for accurate gamut management, however, is likely to increase in the near future given the current consumer display trends towards higher dynamic range and wider gamut. The recent standard behind Ultra HD in particular (ITU-R Rec. BT.2020~\cite{ITU2012}), specifies a considerably larger gamut than the previous widely adopted ITU-R Rec. BT.709~\cite{ITU1998} color gamut, while concurrent proposals are pushing towards defining content at 4000 or even 10000 nits of peak luminance. At the same time, no displays exist that can achieve a full ITU-R Rec. BT.2020 gamut or these luminance levels. Consequently, both tone mapping and gamut management will be necessary to ensure that content is displayed as intended.

\section{Conclusions}
\label{sec:Conclusions}

Typically tone mapping compresses the luminance range of images, while chromatic
information is left untouched or is manually corrected as a post process. On the
other hand, gamut mapping algorithms deal with both luminance and chromatic
information and aim to map images from one gamut to another, but typically deal
with small changes, mostly along the chromatic dimensions. In this paper, we show
how gamut mapping techniques can be extended to the HDR domain, when the
transitions between the input and output gamuts are often very large. We
integrate tone mapping, chroma compression, and gamut management into a single
framework that preserves the appearance of the HDR input, while preventing
unwanted shifts and ensuring that the output image fits within the target gamut
boundaries.

Within our proposed framework, we describe two chroma compression methods, each
best suited for a different part of the imaging pipeline. We have evaluated our
framework using several tone mapping operators and compared its results with
traditional tone and gamut mapping techniques as well as a color correction
formula to show the ability of our approach to overcome their drawbacks. We
also showed that existing tone mapping operators can be naturally integrated
into our framework, which enables the users to choose different operators for
different applications.

\section{Acknowledgments}
This work was partially supported by Ministry of Science and Innovation Subprogramme Ramon y Cajal RYC-2011-09372, TIN2013-47276-C6-1-R from Spanish government,  2014 SGR 1232 from Catalan government and EC "PARTHENOS" (GA no. 654119) from the Italian government.

\ifCLASSOPTIONcaptionsoff
  \newpage
\fi



\bibliographystyle{IEEEtran}
\bibliography{GM}

\end{document}